\documentclass[letterpaper]{article} % DO NOT CHANGE THIS
\usepackage{aaai23}  % DO NOT CHANGE THIS
\usepackage{times}  % DO NOT CHANGE THIS
\usepackage{helvet}  % DO NOT CHANGE THIS
\usepackage{courier}  % DO NOT CHANGE THIS
\usepackage[hyphens]{url}  % DO NOT CHANGE THIS
\usepackage{graphicx} % DO NOT CHANGE THIS
\usepackage{amsmath}
\usepackage{amssymb}
\usepackage{natbib}  % DO NOT CHANGE THIS AND DO NOT ADD ANY OPTIONS TO IT
\usepackage{caption} % DO NOT CHANGE THIS AND DO NOT ADD ANY OPTIONS TO IT
\usepackage[switch]{lineno}
\usepackage[algo2e]{algorithm2e} 
\usepackage{algorithm}
\usepackage{algorithmic}
\usepackage{caption}
\usepackage{subcaption}
\usepackage{booktabs}
\usepackage[dvipsnames]{xcolor}
\usepackage{tabularx,colortbl}
\usepackage[skins]{tcolorbox}
\newtcbox\tp{hbox, on line, colback=LimeGreen, enhanced, frame hidden, boxrule=0pt, 
    top=-2pt, bottom=-2pt, right=-2pt, left=-2pt, sharp corners}
    
\newtcbox\seccol{hbox, on line, colback=Goldenrod, enhanced, frame hidden, boxrule=0pt, 
    top=-2pt, bottom=-2pt, right=-2pt, left=-2pt, sharp corners}
\usepackage{newfloat}
\usepackage{listings}
\DeclareCaptionStyle{ruled}{labelfont=normalfont,labelsep=colon,strut=off} % DO NOT CHANGE THIS
\floatstyle{ruled}
\newfloat{listing}{tb}{lst}{}
\floatname{listing}{Listing}
\title{PDRF: Progressively Deblurring Radiance Field for Fast and Robust Scene Reconstruction from Blurry Images}

\title{PDRF: Progressively Deblurring Radiance Field for Fast and Robust Scene Reconstruction from Blurry Images}
\author {
    Cheng Peng,
    Rama Chellappa
}
\affiliations {
    Johns Hopkins University \\
    cpeng26@jhu.edu
}
\usepackage{bibentry}
\begin{document}

\maketitle
% \twocolumn[{%
% \renewcommand\twocolumn[1][]{#1}%
% \maketitle
% \vspace{-3em}
% \input{figures/title}
% }]

\begin{abstract}
We present Progressively Deblurring Radiance Field (PDRF), a novel approach to efficiently reconstruct high quality radiance fields from blurry images. While current State-of-The-Art (SoTA) scene reconstruction methods achieve photo-realistic rendering results from clean source views, their performances suffer when the source views are affected by blur, which is commonly observed for images in the wild. Previous deblurring methods either do not account for 3D geometry, or are computationally intense. To addresses these issues, PDRF, a progressively deblurring scheme in radiance field modeling, accurately models blur by incorporating 3D scene context. PDRF further uses an efficient importance sampling scheme, which results in fast scene optimization. Specifically, PDRF proposes a Coarse Ray Renderer to quickly estimate voxel density and feature; a Fine Voxel Renderer is then used to achieve high quality ray tracing. We perform extensive experiments and show that PDRF is \textbf{15X faster} than previous SoTA while achieving \textbf{better performance} on both synthetic and real scenes.

%HDRF proposes a Coarse Ray Renderer, which provides fast density and color estimation. Based on these estimations, HDRF uses a Hierarchical Kernel Proposal network and a Fine Voxel Renderer to simultaneously refine the proposed kernel and color.
\end{abstract}
\section{Introduction}

Reconstructing a 3D scene from 2D images is a long-standing research problem with extensive applications in robotics, site model construction, AR/VR, e-commerce, etc. Significant progress has been observed recently with developments in differentiable rendering and implicit function representation, e.g. Neural Radiance Field, or NeRF~\cite{DBLP:conf/eccv/MildenhallSTBRN20}. In this approach, scene radiance is represented as a neural network, i.e. a Multi-Layer Perceptron (MLP), which maps 3D coordinates to their respective density and color. By incorporating the MLP into ray tracing, scene radiance is optimized with a self-supervised photometric loss to satisfy multi-view observations; therefore, only calibrated images are required. This formulation produces photo-realistic scene rendering, models scene geometry, and accounts for view dependent effects.
% The scene radiance is modeled as an implicit function that maps 3D coordinates and view directions to voxel density and color. 
% NeRF models a 3D radiance field as an implicit function, which maps 3D coordinates and view directions to color and density at those locations. 
% This implicit function is parameterized as a Multi-Layer Perceptron (MLP) and trained with a self-supervising photometric loss to satisfy multi-view observations. 
% Under such a formulation, only calibrated images are required to produce photo-realistic scene rendering from novel views. Furthermore, this MLP-based implicit function can also produce the underlying 3D scene geometry and model view dependent effects.

\begin{figure}[!htb]
    \centering
      \includegraphics[width=1\linewidth,height=0.7\linewidth]{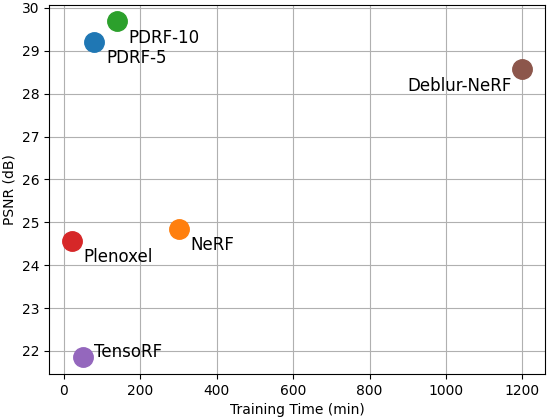}
    \caption{Comparisons of radiance field modeling methods from \emph{blurry} images. Our proposed PDRF significantly outperforms previous methods in speed and performance. Results are aggregated from Table 1.}
    \label{fig:speed_vs_perf}
    \vspace{-1.5em}
\end{figure}

While NeRF, along with other implicit function-based methods~\cite{DBLP:conf/cvpr/ParkFSNL19,DBLP:conf/cvpr/MeschederONNG19,DBLP:conf/cvpr/ChenZ19,DBLP:conf/nips/SitzmannZW19}, marks a significant step towards photo-realistic scene reconstruction, its application in the real world is still limited. Images acquired in the wild are often affected by factors like object motion, camera motion, incorrect focus, or low light conditions. These factors can significantly impact radiance field modeling due to information loss and multi-view \emph{in}consistency. There often still exists enough information within images in the wild to model high quality radiance field, despite the unconstrained acquisition conditions. To this end, subsequent works~\cite{Mildenhall_2022_CVPR,park2021nerfies,park2021hypernerf,DBLP:conf/cvpr/PumarolaCPM21,tretschk2021nonrigid,DBLP:conf/cvpr/Martin-BruallaR21} attempt to account for these conditions with domain knowledge and estimating robust NeRF.
% RawNeRF~\cite{Mildenhall_2022_CVPR} uses an approximated tonemapped loss and variable exposure to account for low light observations. Deblur-NeRF~\cite{ma2022deblur} incorporates a blur model into the rendering pipeline such that the underlying radiance field can be blur-free. These works have shown impressive improvements in radiance field modeling from imperfect observations.

In this work, we seek to better address camera motion and defocus blur in radiance field modeling, which is commonly observed. Prior works have two drawbacks: a lack of 3D context in modeling blur and a time-consuming optimization process. In particular, Deblur-NeRF~\cite{ma2022deblur} models a blurry observation by blending $P$ clean pixels with a blur kernel; however, the pixel-wise blur is predicted based on 2D image coordinates. As we will show, this limits Deblur-NeRF's blur modeling capacity due to the lack of scene context. Since $P$ pixels need to be ray-traced to render a single blurry pixel, Deblur-NeRF is also $P$ times slower than NeRF in rendering the same number of pixels. As shown in Fig. 1, this make Deblur-NeRF very computationally demanding.%, as NeRF is already computationally demanding to optimize.

Several previous works~\cite{fridovich2022plenoxels,mueller2022instant,SunSC22,tensorf} have shown great progress in accelerating NeRF; however, they are not directly applicable to methods that dynamically generated rays to account for view inconsistencies. Specifically, these acceleration methods rely on an explicit density volume, which allows dense ray-tracing by quickly pruning away low density voxels. This strategy relies on the assumption that static rays can lead to good density approximation, which is not true when images are blurred. Voxel pruning also leads to sub-optimal learning for dynamic ray generation because gradients only propagate on parts of the ray. %Due to its limited capacity, an explicit density volume also exhibits significant noise when optimized on images with complex details and blur.

We propose a novel Progressively Deblurring Radiance Field, or PDRF. PDRF uses a Progressive Blur Estimation (PBE) module to address the lack of 3D modeling for blur. To achieve this, PBE first generates a coarse set of rays to account for blur based on 2D coordinates. PBE then traces these rays and obtains their features from the underlying radiance field. Finally, PBE refines coarse blur estimations with a second update by incorporating 3D features into blur modeling. For rendering, PDRF uses a Coarse Ray Renderer (CRR) and a Fine Voxel Renderer (FVR). CRR provides feature and density estimation for PBE and FVR respectively. In particular, CRR achieves efficiency by rendering a ray from its aggregated feature instead of rendering individual voxels. FVR focuses on quality, and uses a larger network and denser predictions for high quality view synthesis. CRR and FVR combine the use of explicit features~\cite{tensorf} and an efficient importance sampling to enable fast convergence and proper gradient updates for PBE. As demonstrated in Fig. 1, despite the more sophisticated blur modeling, PDRF is faster \emph{and} more performant than Deblur-NeRF in modeling radiance field from blurry images. In summary, our contributions can be described in three parts:

% At the core of HDRF is a Coarse Ray Renderer (CRR), which provides fast opacity and ray feature estimations. CRR performs pixel rendering from ray-tracing voxel features instead of voxel-specific rendering. This implementation is 

% . To perform deblurring, HDRF proposes a Hierarchical Kernel Proposal (HKP) network. HKP consists of a coarse and refined stage; specifically, a coarse blur kernel is first proposed for CRR to render; based on the pixel features produced by CRR, HKP further refines its kernel proposal. The refined kernel is then rendered by a combination of CRR and Fine Voxel Renderer (FVR), which takes into consideration the ray tracing direction and renders color for every voxel. By incorporating TensoRF's decomposed volume representation but rendering without online pruning, CRR and HKP is both compatible to ray optimization and extremely fast. 

\begin{enumerate}

\item We propose a Progressive Blur Estimation module, which first proposes a coarse blur model and updates it with corresponding 3D features to capture details. 

\item We propose an efficient rendering scheme, which includes a Coarse Ray Renderer and a Fine Voxel Renderer; this scheme is significantly faster than NeRF while supporting ray optimization in PBE.

\item We evaluate the overall PDRF method on synthetic and real images with blur, and find significant performance improvements compared to SoTA methods. 
\end{enumerate}

%HDRF consists of two components: a Hierarchical Kernel Proposal (HKP) network, and a fast, Coarse-to-Fine Tensorial Radiance Field (C2F-TensoRF). At the core of C2F-TensoRF lies the Coarse Ray-Renderer (CRR), which addresses multiple challenges. Specifically, CRR is very fast at estimating both color and density, which are necessary for HKP and the Fine Voxel Renderer to work. 

% we find that the DSK proposal network in Deblur-NeRF often leads to sub-optimal kernel proposal. Specifically, we observe that the kernel proposals happen before the pixel are rendered, which means that DSK has no information about the pixel values it tries to deblur other than its location. This is in contrast to many previous Deep-Learning based work which 

\section{Related Work and Backgrounds}

\noindent\textbf{Motion and Defocus Deblurring.}
Motion and defocus blur are commonly observed image degradations. Blurry observations are typically modeled as convolutions between clean images and blur kernels; therefore, single image deblurring is a heavily ill-posed problem, as many solutions can lead to the same observations. Traditionally, various priors are used to constraint the solution space~\cite{DBLP:journals/tip/ChanW98,RUDIN1992259,DBLP:conf/cvpr/KrishnanTF11,DBLP:conf/cvpr/XuZJ13,5674049,DBLP:conf/nips/Levin06}. More recently, deep learning-based methods have achieved great performances by learning a more sophisticated deblurring prior in a data driven way, e.g. as a function that directly maps blurry images to clean images.~\cite{DBLP:conf/eccv/Chakrabarti16,DBLP:conf/cvpr/KupynBMMM18,DBLP:conf/cvpr/LeeSRCL21,DBLP:conf/cvpr/NahKL17,9709995,DBLP:conf/cvpr/TaoGSWJ18,DBLP:conf/cvpr/ZamirA0HK0021,DBLP:conf/cvpr/ZhangLZMSLL20} Video-based deblurring explores temporal consistency to find better solutions, e.g. through optical flow~\cite{DBLP:conf/cvpr/KimL15a,DBLP:conf/cvpr/PanBT20,DBLP:conf/cvpr/LiXZ0ZRSL21} or recurrent structures~\cite{DBLP:journals/corr/abs-2106-16028,DBLP:conf/iccv/KimLSH17,DBLP:conf/cvpr/NahSL19,DBLP:journals/tog/SonLLCL21}.

% conventionally, blurry pixels $\tilde{C}$ are modeled by convolution operations $\ast$:
% \vspace{-0.2em}
% \begin{equation}
% \label{Eq:blur}
%     \tilde{C}(x) = C_p(x) \ast h,
% \end{equation}

\noindent\textbf{Neural Radiance Field Modeling.} Significant progress has been made in recent years on the 3D scene reconstruction task. Due to space constraints, we refer to previous surveys~\cite{DBLP:journals/pami/HanLB21,DBLP:journals/cgf/TewariFTSLSMSSN20} for a comprehensive review. We focus on neural radiance field modeling, a SoTA approach for scene reconstruction. NeRF~\cite{DBLP:conf/eccv/MildenhallSTBRN20} is a seminal work that models scene radiance with an implicit MLP function $F$. The color $\textbf{c}_i$ and density $\sigma_i$ at location $X_i\in \mathbb{R}^3$ can be queried from $F$:
\begin{equation}
\label{nerf}
\begin{gathered}
    (\textbf{c}_i,\sigma_i)=F(\gamma_{L_X}(X_i),\gamma_{L_d}(d)),
\end{gathered}
\end{equation}
where $\textbf{c}_i$ is dependent on view direction $d$, and $\gamma_{L}$ is an $L$-frequency-band embedding function that leads to better convergence~\cite{tancik2020fourfeat}. By tracing a ray at locations $X_i=o+t_{i}d$ , where $o$ is the ray origin and $t_{i}$ indicates the travel distance, the rendered color $\hat{C}(\textbf{r}_x)$ for ray $\textbf{r}_x$ is:
\begin{equation}
\label{render}
\begin{gathered}
    \hat{C}(\textbf{r}_x)=\sum_{i=1}^{N} T_i(1-\textrm{exp}(-\sigma_i\delta_i))\textbf{c}_i,\\
    T_i = \textrm{exp}(-\sum_{j=1}^{i-1}\sigma_j\delta_j),
\end{gathered}
\end{equation}
which aggregate color $\textbf{c}_i$ based current and cumulative density $\{\sigma_i, T_i\}$, modulated by distance between samples $\delta_i$. We update $F$ by constraining $\hat{C}(\textbf{r}_x)$ to be similar to observations $C(\textbf{r}_x)$ at all pixel locations $x\in \mathbb{R}^2$.
%$\lVert \hat{C}(\textbf{r}_x)-C(\textbf{r}_x)\rVert^{2}_{2}$ 
% where $\hat{C}(\textbf{r})$ is the rendered pixel from a ray $\textbf{r}$, $\sigma_i$ is the opacity for different points along the ray; $T_i = \textrm{exp}(-\sum_{j=1}^{i-1}\sigma_j\delta_j)$ is the cumulative opacity, $\delta_i=t_{i+1}-t_i$ and $t_i$ are the distance between two traced voxels and the voxel distance along the ray. NeRF estimates $\textbf{c}_i$ and $\sigma_i$ in Eq.~(\ref{render}) by modelling the underlying radiance field with a neural network function $\mathcal{F}$:
% \begin{equation}
% \label{nerf}
% \begin{gathered}
%     (\textbf{c},\sigma)=\mathcal{F}(\gamma_{L_x}(v),\gamma_{L_d}(d)),\\
%     \gamma_{L}(v)=[\textrm{sin}\pi v, \textrm{cos}\pi v, ...,\textrm{sin}2^{L-1}\pi v, \textrm{cos}2^{L-1}\pi v],
% \end{gathered}
% \end{equation}
% where $v$ is the 3D coordinate of a voxel, $d$ is the ray direction, and $\gamma_L(v)$ is the positional embedding function that is essential for better convergence~\cite{tancik2020fourfeat}. 

NeRF can produce photo-realistic renderings on novel views; however, it relies on a powerful MLP to represent the scene. This leads to a very long optimization time as millions of voxel locations need to be queried from a network. To ameliorate this issue, NeRF applies importance sampling: it first samples few equidistant locations along a ray before proposing additional points around high density regions for finer sampling. Many follow-up works seek to further accelerate NeRF optimization, generally following a space-time trade-off by introducing explicit representations to supplement implicit neural networks.
DirectVoxGo~\cite{SunSC22} proposes to use feature and density volumes to store view-invariant information and use a shallow MLP to render view-dependent color. Plenoxel~\cite{fridovich2022plenoxels} discards MLPs and uses spherical harmonics as the function basis to model view dependent effects; the voxel density and spherical harmonic coefficients are stored as volumes. A downside to using volumetric representations is the large memory footprint. To this end, Instant-NGP~\cite{mueller2022instant} uses multi-resolution hashing instead of a volume to compactly encode explicit features. TensoRF~\cite{tensorf} proposes a tensor decomposition to approximate 3D features in lower dimensions, which is more scalable to high resolution.
Another approach to accelerate NeRF is through more efficient importance sampling. Mip-NeRF360~\cite{barron2022mipnerf360} uses a small coarse network that outputs density only and is supervised by another large network. PDRF combines a novel and efficient importance sampling scheme with explicit representations to realize greater acceleration. %In spirit, this is the closest to our proposed CRR; however, CRR estimates both opacity and color to assist blur kernel estimation and also leverages explicit feature representations for speed.

% If the observed $C(r)$ is blurred, the original rendering and photometric loss alone will lead to sub-optimal Radiance Field modelling. The sub-optimalities manifest as either blurry underlying structures or noisy opacity values near the camera plane; therefore, a deblurring formulation is needed.

\noindent\textbf{NeRF Modeling With Non-Ideal Images.}
% Motion and defocus blur are commonly observed image degradations; conventionally, blurry pixels $\tilde{C}$ are modeled by convolution operations $\ast$:
% \vspace{-0.2em}
% \begin{equation}
% \label{Eq:blur}
%     \tilde{C}(x) = C_p(x) \ast h,
% \end{equation}
% where $C_p(x)$ is the clean image patch around pixel location $x$, and $h$ is the blurring kernel. In general, there is no prior knowledge of the type of blur on the image, i.e. blind deblurring, which seeks the inverse model of $h$ such that $C$ can be recovered from $\tilde{C}$. This task is heavily ill-posed but has seen much progress by leverage large-scale, supervised deep learning methods; the inverse of $h$ is often modeled by deep Convolutional Neural Networks (CNNs)~\cite{}. While the results are impressive, supervised learning requires either expensive data collection~\cite{koreanpaperonrealisticblur} or synthetic data and is often lacking in generalization. 
NeRF works well when images are clean and well-calibrated; however, images acquired in the wild are often non-ideal. Factors, such as low light, camera motion, object motion, and incorrect focus, can degrade image quality or affect multi-view consistency. Many works introduce domain knowledge to model the non-ideal contributor to observations, such that the radiance field function only models a canonical, multi-view consistent scene. RawNeRF~\cite{Mildenhall_2022_CVPR} uses an approximated tonemapped loss and variable exposure to account for low light observations. NeRF-W~\cite{DBLP:conf/cvpr/Martin-BruallaR21} uses image-specific embedding to model inconsistent appearances and transient objects. For scenes with moving objects, many works~\cite{park2021nerfies,park2021hypernerf,DBLP:conf/cvpr/PumarolaCPM21,tretschk2021nonrigid} use an implicit function to describe scene deformation, such that the rays are first deformed, then traced by a radiance field function. Recently, Deblur-NeRF~\cite{ma2022deblur} introduces an implicit function to model per-pixel blur before radiance field modeling. Specifically, the blurry observation $\tilde{C}$ is modeled as a convolution between clean pixels and a blur kernel:
% to jointly address radiance field modelling and blurry source views. Specifically, it uses a sparse blur kernel formulation, which models blur as the following:
\vspace{-0.1em}
\begin{equation}
\label{Eq:sparseblur}
    \tilde{C}(\textbf{r}_x) = \sum_{x_i\in \mathcal{N}(x)}C(\textbf{r}_{x_i})h(x_i) \text{ s.t } \sum_{x_i\in \mathcal{N}(x)}h(x_i)=1,% C_p(x) \ast h; C_p, h \in \mathbb{R}^{K\times K,}
\end{equation}
where $x_i$ denotes the clean pixel coordinates at neighboring locations $\mathcal{N}(x)$. The implicit function accounts for blur by predicting $P=|\{x_i\}|$ rays and their blending weights $h(x_i)$, which are used by a NeRF to computes $\tilde{C}(\textbf{r}_{x_i})$. %Due to the additional modeling, this method is more costly to optimize than NeRF; as such, a faster method that is suitable for dynamically updated rays is of significant interest.

\section{Progressively Deblurring Radiance Field}

\begin{figure*}[!htb]
    \centering
      \includegraphics[width=1.0\textwidth]{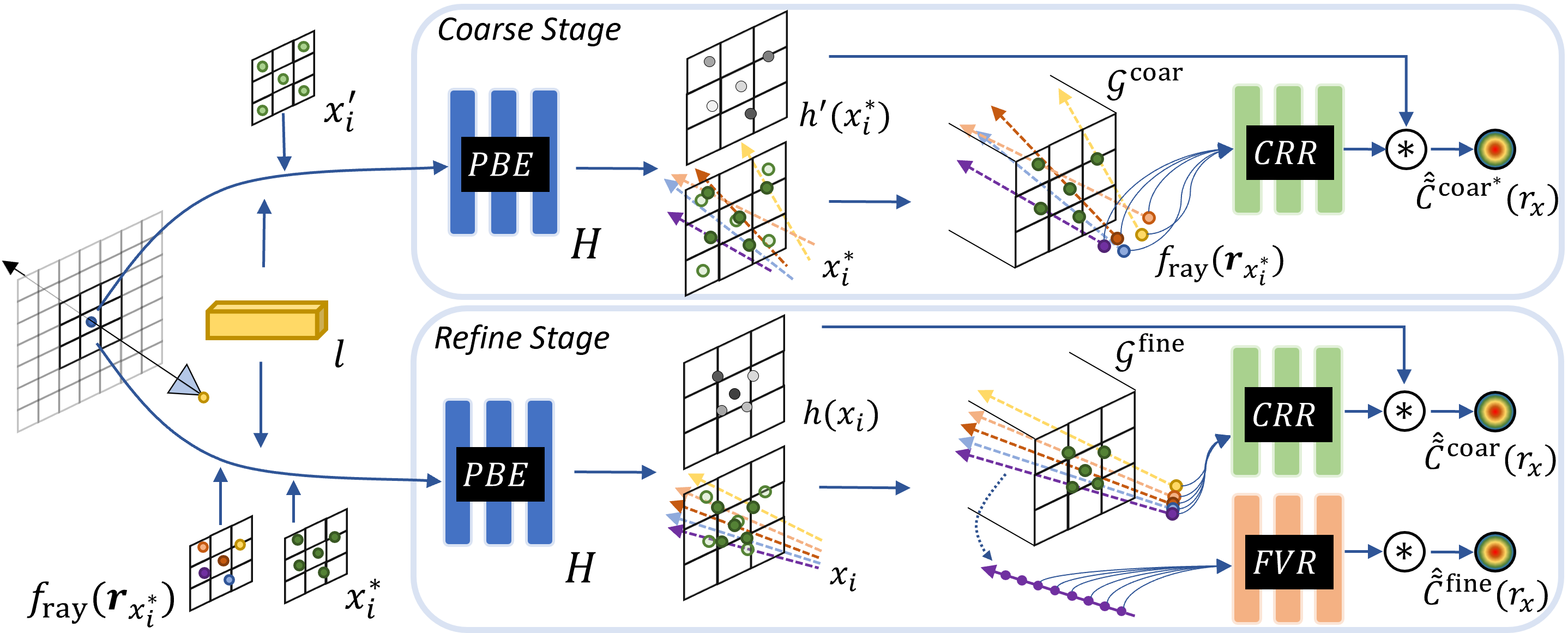}
    \caption{PDRF is split into two stages. In the coarse stage, PBE models the blur from 2D coordinates to obtain the initial kernel locations $x^{*}_i$. We then obtain the respective ray features $f_{\textrm{ray}}(\textbf{r}_{x^{*}_i})$ from CRR. PBE then generates a second, refined blur estimate based on $\{x^{*}_i,f_{\textrm{ray}}(\textbf{r}_{x^{*}_i})\}$, and uses CRR and FVR to render the observed blurry pixel. After optimization, we can directly render from a blur-free radiance field to obtain deblurred images. }
    \label{fig:network}
    \vspace{-0.5em}
\end{figure*}

As shown in Fig.~\ref{fig:network}, PDRF consists of two parts: a Progressive Blur Estimation (PBE) module, and a rendering module that contains a Coarse Ray Renderer (CRR) and a Fine Voxel Renderer (FVR). We first describe PBE, which models blur in two stages. The coarse stage is fast and estimates blur based on 2D pixel coordinates; the finer stage updates previous results by adding ray features, such that PBE can take into account 3D scene context and model blur with more precision. We then describe CRR and FVR, which combine an efficient importance sampling scheme with explicit representations to accelerate optimization. CRR aims to provide fast ray feature and density estimations for PBE and FVR.
%; it achieves so by rendering color from aggregated ray features instead of on individual voxel. 
FVR performs importance sampling based on CRR's density estimation and have finer ray tracing resolution; it also uses a larger, voxel-wise rendering network to produce high quality view synthesis. %By combining these designs, PDRF achieves better performance in deblurring and faster speed. 

%The solution space that determines $\textbf{r}_{x_i}$, i.e. $\{o_{x_i}, d_{x_i}\}\in \mathbb{R}^{N\times P \times 6}$ for all $N$ observed pixels, is large. 

\subsection{Progressive Blur Estimation}
% \noindent\textbf{Progressive Blur Estimation.}
We follow prior works by introducing an implicit function $H$ to model blur. Specifically, $H$ estimates a number of $\textbf{r}_{x_i}$ and their blur kernel weights $h(x_i)$ for every $x$; we then follows Eq. (\ref{Eq:sparseblur}) to render the blurry observations.
%To model radiance field from blurry images, an implicit function is used to estimates $P$ number of $\textbf{r}_{x_i}$ and their blur kernel weights $h(x_i)$, which are rendered and combined following Eq. (\ref{Eq:sparseblur}). 
Since tracing $C(\textbf{r}_{x_i})$ is costly, the kernel size $P$ should be minimal. To effectively find the optimal $P$ rays to describe blur, we propose a two-stage estimation scheme. As shown in Fig. 2, the coarse blur estimation stage makes use of a canonical blur kernel location $x'_i\in \mathbb{R}^{V\times P\times 2}$ and view embedding $l\in \mathbb{R}^{V\times K}$, where $V$ is the number of source views and $K$ is the embedding size. Both $x'_i$ and $l$ are learnable parameters that effectively serve as the blur kernel and view index to $H$. We predict $\textbf{r}_{x_i}$ from $H$ in the following way:

\begin{equation}
\label{Eq:PKP_1}
\begin{gathered}
    \Delta o_{x'_i}, \Delta x'_i, h'(x^*_i) = H(l,x,x'_i,\emptyset), \\
    x^{*}_i = x'_i + \Delta x'_i, o_{x^{*}_i} = o+\Delta o_{x'_i},\\
    \textbf{r}_{x^{*}_i} = o_{x^{*}_i} + td_{x^{*}_i},
\end{gathered}
\end{equation}
where $\emptyset$ is a zero vector placeholder for additional features. This initial blur model $\{\textbf{r}_{x^{*}_i}, h'(x^*_i)\}$ is similar in formulation to Deblur-NeRF~\cite{ma2022deblur} and has been shown to work reasonably well. The implicit function $H$ compactly represents all blur in the scene, and $l,x,x'_i$ are the related indices to query the specific blur model.
% In concept, $H$ is similar to the radiance field function $F$, where $l,x,x'_i$ are the related indices to be queried. As $\textbf{r}_{x^{*}_i}$ are traced and blended, $H$ then fit to the specific scenes to account for blur. 
As recent works~\cite{fridovich2022plenoxels,SunSC22,mueller2022instant,tensorf} suggest that an implicit function can be enhanced by explicit features for better and faster convergence, we hypothesize that $H$ can also benefit from additional features. 

\begin{figure}[!htb]
    \setlength{\abovecaptionskip}{3pt}
    \setlength{\tabcolsep}{2pt}
    \begin{tabular}[b]{cc}
        \begin{subfigure}[b]{.48\linewidth}
            \includegraphics[width=\textwidth]{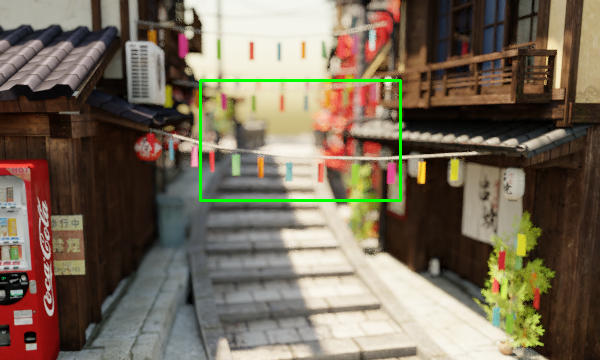}
            \caption{Defocus Blur}
            \label{img:blur}
        \end{subfigure} &
        \begin{subfigure}[b]{.48\linewidth}
            \includegraphics[width=\textwidth]{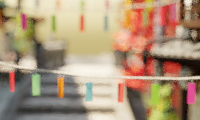}
            \caption{Deblur Blur, Magnified}
            \label{}
        \end{subfigure} \\
        \begin{subfigure}[b]{.48\linewidth}
            \includegraphics[width=\textwidth]{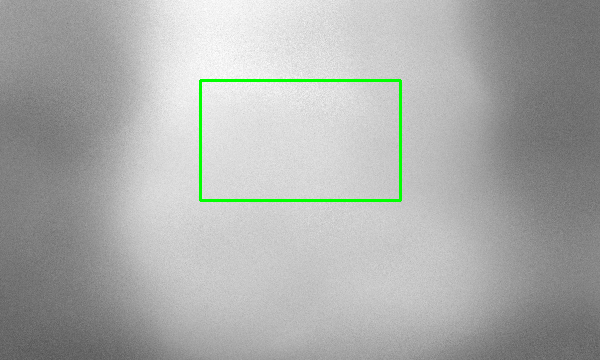}
            \caption{$Var(\Delta o_{x^{*}_i})$, Ma et al.}
            \label{img:DSK}
        \end{subfigure} &
        \begin{subfigure}[b]{.48\linewidth}
            \includegraphics[width=\textwidth]{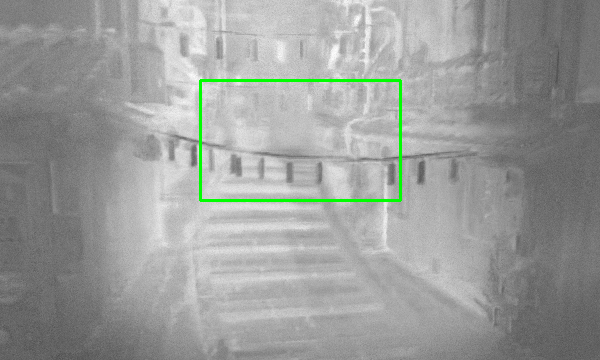}
            \caption{$Var(\Delta o_{x_i})$, PBE}
            \label{img:PRE}
        \end{subfigure} \\
    \end{tabular}
    \caption{Visualization of defocus blur in (a), background is out-of-focus. We visualize the variance of predicted $\Delta o_{x_i}$, where higher variance correlates with more blurry regions. Our PBE module (d) can clearly distinguish foreground details despite their proximity with background pixels; (c) can only coarsely distinguish fore/background.}
    \label{fig:pre}
    \vspace{-0.5em}
\end{figure}

To motivate the correct features that can help $H$ account for blur, consider a complex scene in Fig.~\ref{img:blur} where the background is out-of-focus. A key piece of information that describes blur in a pixel is its depth, as pixels have more blur when their content is further away from the camera. While $\{\textbf{r}_{x^{*}_i}, h'(x^*_i)\}$ can model a coarse blurry regions, it cannot distinguish the sharp foreground details when those details are amongst background pixels, limited by their similar 2D coordinate inputs. 
Based on this observation, we introduce a second, refinement stage that incorporates ray features $f_{\textrm{ray}}$ to $H$ for fine-grained blur modeling. As shown in Fig. 2, $f_{\textrm{ray}}(\textbf{r}_{x^{*}_i})$ is obtained by tracing and aggregating per-voxel features from the underlying radiance field. PBE performs a second inference to update $\{\textbf{r}_{x^{*}_i}, h'(x^*_i)\}$:
% where $H$ takes in an additional ray feature parameter; as we have no knowledge of the ray feature at first, a zero vector $\emptyset$ is used as a placeholder. In the second, radiance field-aware step, we update $\{\Delta o_{x'_i}, \Delta x'_i\}$ by introducing ray features $f_{\textrm{ray}}$ to $H$; the ray displacements are refined as follows:
\begin{equation}
\label{Eq:PKP_2}
\begin{gathered}
    \Delta o_{x^{*}_i}, \Delta x^*_i, h(x_i) = H(l,x,x^*_i,f_{\textrm{ray}}(\textbf{r}_{x^{*}_i})),\\
    x_i = x^*_i + \Delta x^*_i, o_{x_i} = o+\Delta o_{x^*_i},\\
    \textbf{r}_{x_i} = o_{x_i} + td_{x_i}.
\end{gathered}
\end{equation}
% where $t$ is the ray tracing step size and $d$ is the ray direction.

The updated $\{\textbf{r}_{x_i}, h'(x_i)\}$ is more accurate in describing blur, which we demonstrate in Fig.~\ref{img:DSK} and \ref{img:PRE} by measuring the variances of $\Delta o_{x^{*}_i}$ and $\Delta o_{x_i}$. When observations are already sharp, i.e. a single ray can approximate the observation, $\{\Delta o, \Delta x'\}$ should converge to similar spatial locations; when observations are blurry, $\{\Delta o, \Delta x'\}$ should be dissimilar to account for blur. Clearly, $\{\textbf{r}_{x_i}, h'(x_i)\}$ can better distinguish the clear foreground details from the blurry background.%; furthermore, edges in the blurry background also have more variance. 
While we can alternatively use a single stage to estimate $\{\Delta o, \Delta x'\}$ by replacing $\emptyset$ with $f_{\textrm{ray}}(\textbf{r}_{x^{'}_i})$ in Eq. (4), a two-stage design is still efficient since Eq. (4) computes quickly. The two-stage design also benefits from a more accurate $x^{*}_i$, which allows a more informative $f_{\textrm{ray}}(\textbf{r}_{x^{*}_i})$ .

\vspace{-0.5em}
\subsection{Coarse Ray Renderer}
% \noindent\textbf{Coarse Ray Renderer}
The Coarse Ray Renderer serves two roles in PDRF: it provides coarse density estimation to FVR and ray features to PBE. CRR accomplishes this by performing a fast ray tracing.
To estimate voxel density, CRR uses a two-layer MLP $\mathcal{M}^{\textrm{coar}}_\sigma$ and an explicit feature volume $\mathcal{G}^{\textrm{coar}}$: 
\begin{equation}
\label{Eq:CRR}
    \sigma^{\textrm{coar}}_i, f^{\textrm{coar}}_{X_i} = \mathcal{M}^{\textrm{coar}}_\sigma(\gamma_{L_X}(X_i), \mathcal{G}^{\textrm{coar}}(X_i)),
\end{equation}
where $X_i$ is the traced 3D coordinates. We implement $\mathcal{G}^{\textrm{coar}}$ with tensor decomposition~\cite{tensorf} and use it to accelerate convergence for $\mathcal{M}^{\textrm{coar}}_\sigma$. 
%This design is different from previous acceleration works~\cite{fridovich2022plenoxels,SunSC22,tensorf}, which use explicit density volumes for faster access. 
We find that a purely explicit density representation used by previous methods~\cite{fridovich2022plenoxels,SunSC22,tensorf} is prone to significant noise when the observations have complex details or blur, likely due to its inflexible volumetric data structure. Therefore, Eq.~(\ref{Eq:CRR}) uses a combination of an implicit function and explicit features to ensure good density estimation. CRR further obtains an efficient ray feature representation by modifying on Eq.~(\ref{render}):
% and $\mathcal{G}^{\textrm{coar}}_f(X_i)$ is its feature as defined in Eq. (3). We further propose an efficient ray feature representation by making a modification on Eq.~(\ref{render}):
% We note that, coarse opacity estimation is a common strategy used in radiance field modelling to leverage the natural sparsity in scenes.  Some works~\cite{barron2022mipnerf360,DBLP:conf/eccv/MildenhallSTBRN20} use a dedicated MLP to estimate coarse opacity; due to the longer query time for opacity from an implicit representation, these methods estimate few voxel points first and perform importance sampling around high opacity regions for fine rendering; others~\cite{fridovich2022plenoxels,SunSC22,tensorf} use an explicit voxel grid to store opacity values, which can be accessed in fine resolution and prune off quickly on-the-fly. As online pruning leads to optimization issues for kernel proposal, importance sampling is desired. While explicit opacity representation still has a speed advantage over implicit representation, as shown in Fig. X, the speed advantage in explicit representation comes with an disadvantage in robustness to blurring noise, manifested as noisy opacity near the camera. As such, CRR adopts a dedicated MLP for opacity estimation. To render color quickly, CRR makes a modification on the formulation in Eq.~(\ref{render}) to:
\vspace{-1em}
\begin{equation}
\label{render2}
    f_{\textrm{ray}}(\textbf{r}_x)=\sum_{i=1}^{N_c} T^{\textrm{coar}}_i(1-\textrm{exp}(-\sigma^{\textrm{coar}}_i\delta_i))f^{\textrm{coar}}_{X_i},
\end{equation}
% \vspace{-0.5em}
where $f_{\textrm{ray}}(\textbf{r}_x)$ aggregates feature along ray $\textbf{r}_x$, similar to the aggregated color $C(\textbf{r}_x)$, and $N_c$ is the number of equidistant samples. We find $f_{\textrm{ray}}(\textbf{r}_x)$ to be an effective 3D representation for PBE; in particular, $f_{\textrm{ray}}(\textbf{r}_x)$ is agnostic to ray direction as the blur model in PBE is already view specific. We use $f_{\textrm{ray}}(\textbf{r}_x)$ for color rendering by another MLP $\mathcal{M}^{\textrm{coar}}_r$:
\vspace{-0.5em}
\begin{equation}
\label{render3}
    \hat{C}^{\textrm{coar}}(\textbf{r}_x)= \mathcal{M}^{\textrm{coar}}_r(f_{\textrm{ray}}(\textbf{r}_x), \gamma_{L_d}(d)).
\end{equation}
% where $\gamma_{L_d}(d)$ provides the ray direction encoding of $\textbf{r}$. 
%We find this color rendering formulation to be effective and efficient, despite the lack of per-pixel color rendering that incorporates ray direction. 
%which are of complexity $\mathcal{O}(BN)$, where $B$ is the number of ray and $N$ is the traced voxels per ray
This ray-rendering formulation has two advantages. First, it is very fast to compute; instead of querying every voxel along the ray through an MLP,  Eq.~(\ref{render3}) is only inferenced once per ray. It also serves as a coarse approximation of per-voxel rendering. Consider ray tracing when the surface density is a Dirac delta function $\sigma=\hat{\delta}(X_{\textrm{surf}})$ at a 3D location $X_{\textrm{surf}}$; this simplifies Eq. (7) and Eq. (8) to $\hat{C}^{\textrm{coar}}(\textbf{r})= \mathcal{M}^{\textrm{coar}}_r(f^{\textrm{coar}}_{X_{\textrm{surf}}}, \gamma_{L_d}(d))$, which is equivalent to per-voxel rendering. While this is the optimal solution, multi-view observations constrain $f_{\textrm{ray}}$ to coarsely represent surface features and help CRR achieve a good balance between speed and performance. CRR is first used by PBE to gather ray features and renders color based on $\{\textbf{r}_{x^*_i}, h'(x^*_i)\}$; it is then used with FVR to render colors based on the updated $\{\textbf{r}_{x_i}, h(x_i)\}$. Following Eq. (3), the predicted blurry colors are computed as follows:
\begin{equation}
\begin{gathered}
\hat{\tilde{C}}^{{\textrm{coar}}^*}(\textbf{r}_x) = \sum_{x^*_i} \hat{C}^{\textrm{coar}}(\textbf{r}_{x^*_i})h'(x^*_i), \\\hat{\tilde{C}}^{\textrm{coar}}(\textbf{r}_x) = \sum_{x_i} \hat{C}^{\textrm{coar}}(\textbf{r}_{x_i})h(x_i).
\end{gathered}
\end{equation}
Its training loss is a combined photometric loss with all blurry observation $\tilde{C}(\textbf{r}_x)$:
\begin{equation}
\mathcal{L}^{CRR}_{photo} = \lVert \hat{\tilde{C}}^{{\textrm{coar}}^*}(\textbf{r}_x)-\tilde{C}(\textbf{r}_x)\rVert^{2}_{2} + \lVert \hat{\tilde{C}}^{\textrm{coar}}(\textbf{r}_x)-\tilde{C}(\textbf{r}_x)\rVert^{2}_{2}.
\end{equation}
%In fact, we also obtain competitive results in terms of performance and speed by using Eq. (11) and Eq. (12) with TensoRF~\cite{tensorf}; demonstrations can be found in the supplemental material section.

% While this simplified formulation loses some resolution in modelling per-voxel view dependent effects, it is not completely view independent like the coarse stage in~\cite{SunSC22}; in fact, view directional effects are still reasonably modelled, as shown in Fig. Y, demonstrating the robustness of $f_{\textrm{ray}}(\textbf{r}$.

%Secondly, it doubles as an efficient pixel features to be directly transfer to HPK. 

% \noindent\textbf{Fine Voxel Renderer}
\subsection{Fine Voxel Renderer}
% While CRR is lightweight and quickly estimate $f_{\textrm{ray}}(\textbf{r}_{x_i})$ for PBE. 
The Fine Voxel Renderer leverages the density estimation from CRR and generates high quality rendering. CRR calculates weights $w_i=T_i(1-\textrm{exp}(-\sigma^{\textrm{coar}}_i\delta_i))$ for $N_c$ equidistant points in Eq. (\ref{render2}), these weights are normalized to produce a PDF of ray density. FVR uses this PDF to sample additional $N_f$ points on high density regions for finer tracing. Importance sampling scheme is used instead of voxel pruning, as voxel pruning introduces optimization issues for PBE. Specifically, previous works~\cite{fridovich2022plenoxels,SunSC22,tensorf} use explicit density volume and quickly prune off many low density voxels. While this works for \emph{static} rays, PBE dynamically generates rays $\textbf{r}_{x_i}$ based on gradients from the photometric loss. In this case, voxel pruning also eliminates gradients on many points and forces PBE to optimize $\textbf{r}_{x_i}$ base on small segments of the ray. Furthermore, high density voxels are often not correctly estimated from blurry observations and can lead to erroneous pruning. 
% When the rays are dynamically generated by PBE, voxel pruning leads to sub-optimal learning. More specifically, PBE needs to propose multiple rays to model the blur; at the same time, gradients only back-propagate from high opacity voxels due to voxel pruning. However, high opacity voxels are not reliably estimated due to the presence of blur.

% While NeRF~\cite{Mildenhall_2022_CVPR} also has an importance samp

After importance sampling, FVR uses a large MLP, which contains density and color components $\mathcal{M}^{\textrm{fine}}_\sigma$ and $\mathcal{M}^{\textrm{fine}}_c$ respectively for high quality view synthesis. The voxel density $\sigma^{\textrm{fine}}$ is obtained as follows:
\begin{equation}
    \sigma^{\textrm{fine}}, f^{\textrm{fine}}_{X_i} = \mathcal{M}^{\textrm{fine}}_\sigma(\gamma_{L_v}(X_i), \mathcal{G}^{\textrm{coar}}(X_i)\oplus \mathcal{G}^{\textrm{fine}}(X_i)),
\end{equation}
where $\oplus$ is the concatenation operation. We use an additional fine feature volume $\mathcal{G}^{\textrm{fine}}$ with the previous $\mathcal{G}^{\textrm{coar}}$ to maximize explicit feature capacity. Then, per-voxel color is rendered and aggregated as follows:
\begin{equation}
\begin{gathered} 
\textbf{c}_i = \mathcal{M}^{\textrm{fine}}_c(f^{\textrm{fine}}_{X_i}, \gamma_{L_d}(d)),\\
    \hat{C}^{\textrm{fine}}(\textbf{r}_x)=\sum_{i=1}^{N_c+N_f} T^{\textrm{fine}}_i(1-\textrm{exp}(-\sigma^{\textrm{fine}}_i\delta_i))\textbf{c}_i.
\end{gathered}
\end{equation}
A photometric loss is used for FVR, similar to CRR:
\begin{equation}
\begin{gathered}
\hat{\tilde{C}}^{\textrm{fine}}(\textbf{r}_x) = \sum_{x_i} \hat{C}^{\textrm{fine}}(\textbf{r}_{x_i})h(x_i),\\
\mathcal{L}^{FVR}_{photo} = \lVert \hat{\tilde{C}}^{\textrm{fine}}(\textbf{r}_x)-\tilde{C}(\textbf{r}_x)\rVert^{2}_{2}.
\end{gathered}
\end{equation}
Finally, the overall loss includes a total variation constraint  $\mathcal{L}^{*}_{TV}$  on the explicit feature volumes to enforce smoothness, following previous works~\cite{fridovich2022plenoxels,SunSC22,tensorf}:
\begin{equation}
\mathcal{L}_{tot} = \mathcal{L}^{CRR}_{photo} + \mathcal{L}^{FVR}_{photo} + \lambda (\mathcal{L}^{\mathcal{G}^{\textrm{coar}}}_{TV} + \mathcal{L}^{\mathcal{G}^{\textrm{fine}}}_{TV}).
\end{equation}

\begin{table*}[!htb]
    \setlength{\tabcolsep}{3pt}
    % \small
    \centering
    \begin{tabular}[b]{c}
        \begin{tabular}{l|cc|cc|cc|cc|cc|cc|c}
        \toprule
            &\multicolumn{2}{c|}{\textsl{Factory}}&\multicolumn{2}{c|}{\textsl{Cozyroom}}&\multicolumn{2}{c|}{\textsl{Pool}}&\multicolumn{2}{c|}{\textsl{Tanabata}}&\multicolumn{2}{c|}{\textsl{Trolley}}&\multicolumn{2}{c|}{\textsl{Average}}& Time\\
        \midrule
          Camera Motion& \small{PSNR} & \small{SSIM}  & \small{PSNR} & \small{SSIM} & \small{PSNR} & \small{SSIM} & \small{PSNR} & \small{SSIM} & \small{PSNR} & \small{SSIM} & \small{PSNR} & \small{SSIM}  & Hrs\\
        \midrule
        NeRF & 19.32 & 0.4563 & 25.66 & 0.7941  & 30.45 & 0.8354 & 22.22 & 0.6807 & 21.25 & 0.6370 & 23.78 & 0.6807 & 5.00 \\
        
        Plenoxel & 18.42 & 0.4854 & 25.56 & 0.8480 & 30.13 & 0.8481 & 21.67 & 0.6992 & 20.81 & 0.6759 & 23.31 & 0.7113 & 0.42 \\
        % DVGO & 13.52 &  &  & 15.79 &  &  & 20.61 &  & & 12.99 &  &  & 14.00 &  &  & 15.38 &  & \\
        TensoRF & 18.85 & 0.4981 & 24.91 & 0.7975 & 15.89 & 0.3783 & 21.06 & 0.6762  & 20.01 & 0.6584 & 20.53 & 0.6017 & 0.83\\
        
        \midrule
        MPR + NeRF &21.70  & 0.6153 &27.88  & 0.8502 &30.64  & 0.8385 &22.71  & 0.7199  &22.64  & 0.7141 &25.11  & 0.7476 & 5.00\\
        % PVD + NeRF &20.33  & 0.5386  & 0.3667 &27.74  & 0.8296  & 0.1451& 27.56  & 0.7626  & 0.2148 &23.44  & 0.7293  & 0.2542& 23.81& 0.7351& 0.2567 &24.58  & 0.7190  & 0.2475&5.00\\
        Deblur-NeRF & 25.60 & 0.7750 &  \tp{32.08} & 0.9261 &  \tp{31.61} & \seccol{0.8682} & 27.11 & 0.8640  & 27.45 & 0.8632 & 28.77 & 0.8593 & 20.0\\
        
        Deblur-TensoRF & 23.20 & 0.6560 & 30.18 & 0.8890 & 29.83 & 0.7956 & 25.13 & 0.7891 & 25.10 & 0.7894 & 26.69 & 0.7838 & 1.50\\
        $\textrm{Deblur-TensoRF}^{\textrm{pru}}$ & 20.02 & 0.4889 & 28.00 & 0.8570 & 29.66 & 0.7968 & 22.87 & 0.7029 & 22.99 & 0.7231 & 24.71 & 0.7137 & 1.25\\
        \midrule

        $\textrm{PDRF}^{\textrm{single}}$-5 & 25.65 & 0.7793 & 31.37 & 0.9255 & 31.28 & 0.8623 & 27.55 & 0.8809 & 27.43 & 0.8769 & 28.66 & 0.8650 & \tp{1.15}\\

        % $\textrm{PDRF}^{\textrm{single}}$-5 & 25.68 &  &  & 31.68 &  &  & 31.57 &  &  & 27.37 &  &  & 27.43 &  &  & 28.74 &  & \\
        PDRF-5 & \seccol{25.60} & \seccol{0.7786} & \seccol{32.01} & \seccol{0.9310} & \seccol{31.53} & \tp{0.8686}  & \seccol{27.70} & \seccol{0.8851} & \seccol{27.90} & \seccol{0.8841} & \seccol{28.95} & \seccol{0.8695} &\seccol{1.33}\\
        PDRF-10 & \tp{26.56} & \tp{0.8102} & 31.90 & \tp{0.9321} & 31.29 & 0.8657 & \tp{28.21} & \tp{0.8952} & \tp{28.48} & \tp{0.8956} & \tp{29.29} & \tp{0.8798} &2.33\\

        \bottomrule
        \toprule
         Defocus & \small{PSNR} & \small{SSIM}  & \small{PSNR} & \small{SSIM} & \small{PSNR} & \small{SSIM} & \small{PSNR} & \small{SSIM} & \small{PSNR} & \small{SSIM} & \small{PSNR} & \small{SSIM}  & Hrs\\
        \midrule
        NeRF & 25.36  & 0.7847 &30.03  & 0.8926&  27.77  & 0.7266 &23.80  & 0.7811 &22.67  & 0.7103 &25.93  & 0.7791 & 5.00\\
        
        Plenoxel & 25.51 & 0.8579 & 30.33 & 0.9315 & 27.33 & 0.7382 & 23.59 & 0.8371  & 22.42 & 0.7862 & 25.84 & 0.8302 & 0.42 \\
        % DVGO & 13.54 &  &  & 15.82 &  &  & 20.54 &  & & 12.99 &  &  & 13.85 &  &  & 15.35 &  & \\
        TensoRF & 25.08 & 0.8415 & 29.77 & 0.9154 & 16.23 & 0.4112 & 22.97 & 0.8193  & 22.40 & 0.7819  & 23.17 & 0.7539 & 0.83 \\
        
        \midrule
        KPAC + NeRF &26.40  & 0.8194 &28.15  & 0.8592 &26.69  & 0.6589 &24.81  & 0.8147  &23.42  & 0.7495 &25.89  & 0.7803 &5.00\\
        Deblur-NeRF & 28.03  & 0.8628 &31.85& 0.9175& 30.52  & 0.8246& 26.25  & 0.8517  &25.18  & 0.8067  &28.37  & 0.8527 & 20.0\\
        
        Deblur-TensoRF & 27.01 & 0.8257  & 30.39 & 0.8882 & 27.41 & 0.6841 & 24.36 & 0.7835 & 23.91 & 0.7593 & 26.62 & 0.7882  & 1.50\\    
        $\textrm{Deblur-TensoRF}^{\textrm{pru}}$ & 25.55 & 0.7932 & 28.64 & 0.8777 & 27.60 & 0.6954 & 23.66 & 0.7711 & 22.51 & 0.7195 & 25.59 & 0.7714 & 1.25\\
        \midrule    
        
        $\textrm{PDRF}^{\textrm{single}}$-5 & 28.33 & 0.8743 & 31.78 & 0.9222 & 30.43 & 0.8233& 26.26 & 0.8649  & 25.47 & 0.8261 & 28.45 & 0.8622 &\tp{1.15}\\        
        
        PDRF-5 & \seccol{30.34} & \seccol{0.9032} & \seccol{32.10} & \seccol{0.9269} & \seccol{30.48} & \seccol{0.8262} & \seccol{27.31} & \seccol{0.8818} & \seccol{27.05} & \seccol{0.8581} &  \seccol{29.46} & \seccol{0.8792} &\seccol{1.33}\\
        PDRF-10 & \tp{30.90} & \tp{0.9138} & \tp{32.29} & \tp{0.9305} & \tp{30.97} & \tp{0.8408} & \tp{28.18} & \tp{0.9006} & \tp{28.07} &  \tp{0.8799} & \tp{30.08} & \tp{0.8931} &2.33\\
        \bottomrule
        
        \end{tabular}\\
    \end{tabular}        
    \caption{Quantitative comparisons of different radiance field modeling methods on five synthetic blurry scenes. We color code \tp{best} and \seccol{second best} performances.}
    \label{tab:results}
    \vspace{-1.5em}
\end{table*}

\section{Experiments}

\textbf{Dataset.}
We evaluate our deblurring method on the dataset provided by \cite{ma2022deblur}, which contains five synthetic scenes and twenty real world scenes. The synthetic scenes are affected by camera motion blur and defocus blur separately and have blur-free novel view groundtruth. For real world scenes, ten of them are affected by camera motion blur and the remaining ten are affected by defocus blur. While real world scenes have blur-free references, the calibration and exposure may not be consistent with source views. As such, we report quantitative evaluations, specifically PSNR and SSIM, on synthetic scenes and provide quantitative metrics on real world scenes in supplemental material.

\noindent\textbf{Implementation Details.}
We construct PBE as a four-layer MLP with a channel size of 64 and ReLU as the activation function. The size of view embedding $l$ and ray feature $f_\textrm{ray}$ is 32 and 15. The canonical kernel locations $x'_i$ are normalized to be at most $\pm10$ pixels from $x$. For CRR, we use a decomposed feature tensor $\mathcal{G}^{\textrm{coar}}$ that represents 17 million voxels; for FVR, $\mathcal{G}^{\textrm{fine}}$ represents 134 million voxels. The channel dimension for the decomposed $\{X,Y,Z\}$ axes are $\{64,16,16\}$. For CRR, the two-layer MLP $\mathcal{M}^{\textrm{coar}}_{\sigma}$ and three-layer MLP $\mathcal{M}^{\textrm{coar}}_{r}$ have a channel size of 64 for hidden layers. The two-layer MLP $\mathcal{M}^{\textrm{fine}}_{\sigma}$ and three-layer MLP $\mathcal{M}^{\textrm{fine}}_{c}$ in FVR have a channel size of 256 for hidden layers. Overall, the model size for a single scene is 160MB in PDRF, which is smaller than TensoRF~\cite{tensorf}'s 200MB used for forward-facing scenes. PDRF is trained with a batch size of 1024 rays, each ray has coarse samples $N_c=64$ and importance sampled for $N_f=64$ additional voxels. A gamma correction function of $\hat{C}(\textbf{r}_{x})^{\frac{1}{2.2}}$ is used to account for the camera response function, following Deblur-NeRF. Each scene is trained for 25k iterations. Deblur-TensoRF, a baseline that replaces NeRF with TensoRF in Deblur-NeRF, uses explicit density and feature volumes that represent 134 million voxels and traces 160 samples per ray without pruning. Deblur-TensoRF uses a five-layer MLP for rendering, and a channel size of 128 for hidden layers. All time measurements are reported based on one NVIDIA A5000 GPU. 

%& $\textrm{XraySyn}_{ref}$ 
\captionsetup[subfigure]{labelformat=empty}
\begin{figure*}[htb!]
    \setlength{\abovecaptionskip}{3pt}
    \setlength{\tabcolsep}{0.5pt}
    \centering
    \small
    \begin{tabular}[b]{ccccccccc}
        % \hline
         Novel View & Src. View & NeRF & Plenoxel & TensoRF & Db-Ten.RF &Db-NeRF& PDRF & G.T.\\
        & & 5 hrs & 25 mins & 50 mins & 90 mins & 20 hrs& 80 mins& \\
        % \begin{subfigure}[b]{0.097\linewidth}
        %     \includegraphics[width=\textwidth]{figures/imgs/motion/002_NeRF_crop.png}
        %     \caption{}
        % \end{subfigure} &
        % \begin{subfigure}[b]{0.097\linewidth}
        %     \includegraphics[width=\textwidth]{figures/imgs/motion/002_Plenoxel_crop.png}
        %     \caption{}
        % \end{subfigure} &
        % \begin{subfigure}[b]{0.097\linewidth}
        %     \includegraphics[width=\textwidth]{figures/imgs/motion/002_TensoRF_crop.png}
        %     \caption{}
        % \end{subfigure} &
        % \begin{subfigure}[b]{0.097\linewidth}
        %     \includegraphics[width=\textwidth]{figures/imgs/motion/002_DeblurNeRF_crop.png}
        %     \caption{}
        % \end{subfigure} &   
        % \begin{subfigure}[b]{0.097\linewidth}
        %     \includegraphics[width=\textwidth]{figures/imgs/motion/002_Ours_crop.png}
        %     \caption{}
        % \end{subfigure} &  
        % \begin{subfigure}[b]{0.097\linewidth}
        %     \includegraphics[width=\textwidth]{figures/imgs/motion/002_GT_crop.png}
        %     \caption{}
        % \end{subfigure} &  
        % \begin{subfigure}[b]{0.097\linewidth}
        %     \includegraphics[width=\textwidth]{figures/imgs/motion/002_GT_marked.png}
        %     \caption{}
        % \end{subfigure} \\

        \begin{subfigure}[b]{0.20\linewidth}
            \includegraphics[width=\textwidth]{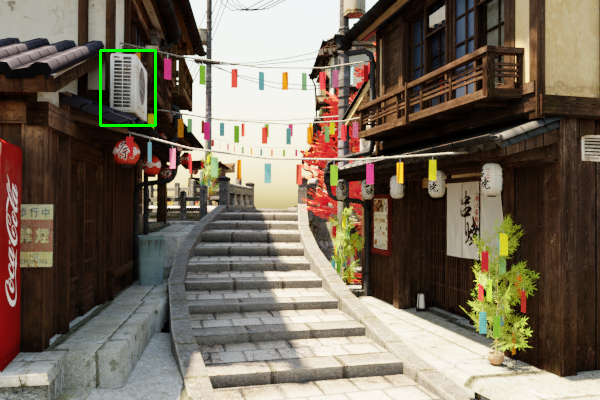}
            \caption{PSNR/SSIM}
        \end{subfigure} &
        \begin{subfigure}[b]{0.097\linewidth}
            \includegraphics[width=\textwidth]{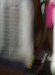}
            \caption{Motion}
        \end{subfigure} &  
        \begin{subfigure}[b]{0.097\linewidth}
            \includegraphics[width=\textwidth]{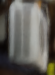}
            \caption{18.98/.538}
        \end{subfigure} &
        \begin{subfigure}[b]{0.097\linewidth}
            \includegraphics[width=\textwidth]{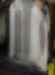}
            \caption{18.51/.530}
        \end{subfigure} &
        \begin{subfigure}[b]{0.097\linewidth}
            \includegraphics[width=\textwidth]{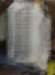}
            \caption{16.11/.321}
        \end{subfigure} &
        \begin{subfigure}[b]{0.097\linewidth}
            \includegraphics[width=\textwidth]{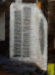}
            \caption{22.00/.726}
        \end{subfigure} &
        \begin{subfigure}[b]{0.097\linewidth}
            \includegraphics[width=\textwidth]{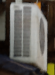}
            \caption{\seccol{23.06/.805}}
        \end{subfigure} &   
        \begin{subfigure}[b]{0.097\linewidth}
            \includegraphics[width=\textwidth]{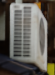}
            \caption{\tp{24.66/.870}}
        \end{subfigure} &  
        \begin{subfigure}[b]{0.097\linewidth}
            \includegraphics[width=\textwidth]{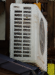}
            \caption{}
        \end{subfigure} \\

        \begin{subfigure}[b]{0.20\linewidth}
            \includegraphics[width=\textwidth]{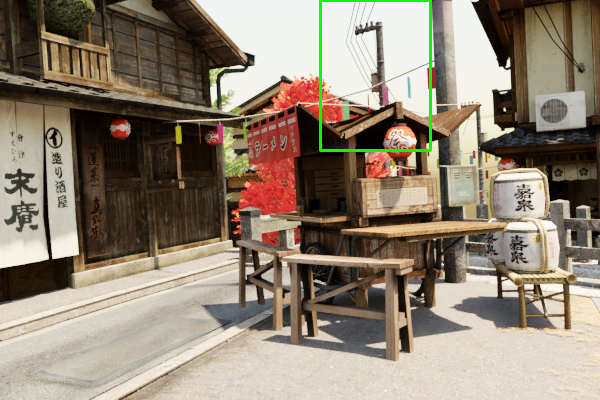}
            \caption{PSNR/SSIM}
        \end{subfigure} &
        \begin{subfigure}[b]{0.097\linewidth}
            \includegraphics[width=\textwidth]{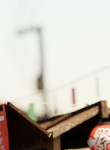}
            \caption{Defocus}
        \end{subfigure} &  
        \begin{subfigure}[b]{0.097\linewidth}
            \includegraphics[width=\textwidth]{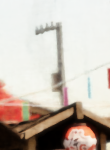}
            \caption{21.71/.798}
        \end{subfigure} &
        \begin{subfigure}[b]{0.097\linewidth}
            \includegraphics[width=\textwidth]{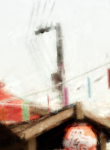}
            \caption{20.80/.778}
        \end{subfigure} &
        \begin{subfigure}[b]{0.097\linewidth}
            \includegraphics[width=\textwidth]{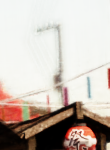}
            \caption{19.33/.758}
        \end{subfigure} &
        \begin{subfigure}[b]{0.097\linewidth}
            \includegraphics[width=\textwidth]{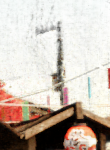}
            \caption{19.33/.758}
        \end{subfigure} &
        \begin{subfigure}[b]{0.097\linewidth}
            \includegraphics[width=\textwidth]{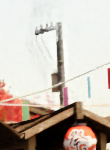}
            \caption{\seccol{23.91/.869}}
        \end{subfigure} &   
        \begin{subfigure}[b]{0.097\linewidth}
            \includegraphics[width=\textwidth]{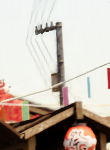}
            \caption{\tp{27.92/.928}}
        \end{subfigure} &  
        \begin{subfigure}[b]{0.097\linewidth}
            \includegraphics[width=\textwidth]{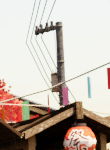}
            \caption{}
        \end{subfigure} \\

        \begin{subfigure}[b]{0.20\linewidth}
            \includegraphics[width=\textwidth]{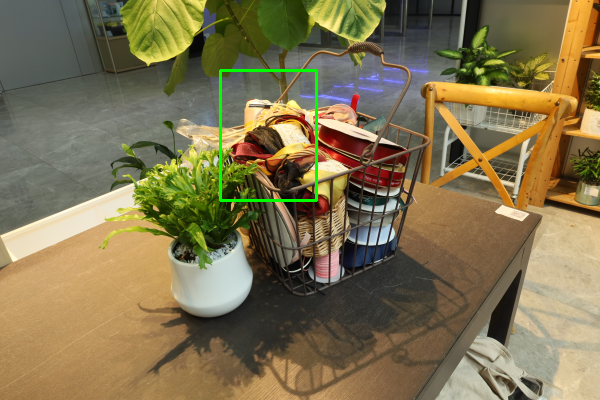}
            \caption{LPIPS}
        \end{subfigure} &
        \begin{subfigure}[b]{0.097\linewidth}
            \includegraphics[width=\textwidth]{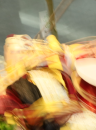}
            \caption{Motion}
        \end{subfigure} &  
        \begin{subfigure}[b]{0.097\linewidth}
            \includegraphics[width=\textwidth]{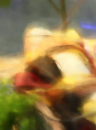}
            \caption{0.2288}
        \end{subfigure} &
        \begin{subfigure}[b]{0.097\linewidth}
            \includegraphics[width=\textwidth]{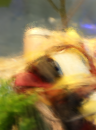}
            \caption{0.2248}
        \end{subfigure} &
        \begin{subfigure}[b]{0.097\linewidth}
            \includegraphics[width=\textwidth]{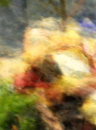}
            \caption{0.2722}
        \end{subfigure} &
        \begin{subfigure}[b]{0.097\linewidth}
            \includegraphics[width=\textwidth]{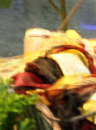}
            \caption{0.1231}
        \end{subfigure} &
        \begin{subfigure}[b]{0.097\linewidth}
            \includegraphics[width=\textwidth]{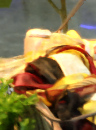}
            \caption{\seccol{0.0953}}
        \end{subfigure} &   
        \begin{subfigure}[b]{0.097\linewidth}
            \includegraphics[width=\textwidth]{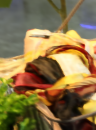}
            \caption{\tp{0.0767}}
        \end{subfigure} &  
        \begin{subfigure}[b]{0.097\linewidth}
            \includegraphics[width=\textwidth]{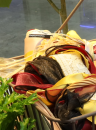}
            \caption{}
        \end{subfigure} \\

        \begin{subfigure}[b]{0.20\linewidth}
            \includegraphics[width=\textwidth]{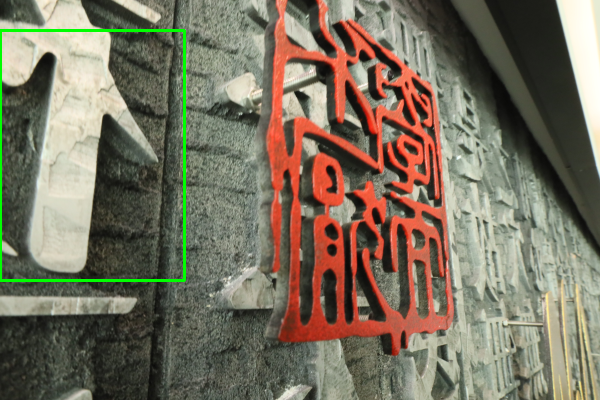}
            \caption{LPIPS}
        \end{subfigure} &
        \begin{subfigure}[b]{0.097\linewidth}
            \includegraphics[width=\textwidth]{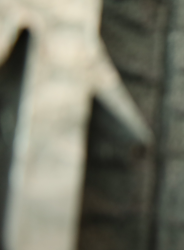}
            \caption{Defocus}
        \end{subfigure} &  
        \begin{subfigure}[b]{0.097\linewidth}
            \includegraphics[width=\textwidth]{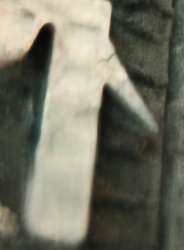}
            \caption{0.1219}
        \end{subfigure} &
        \begin{subfigure}[b]{0.097\linewidth}
            \includegraphics[width=\textwidth]{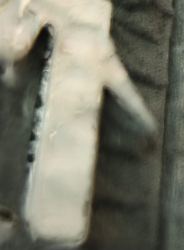}
            \caption{0.1582}
        \end{subfigure} &
        \begin{subfigure}[b]{0.097\linewidth}
            \includegraphics[width=\textwidth]{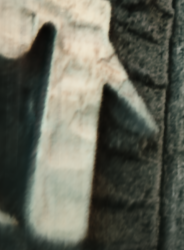}
            \caption{0.1119}
        \end{subfigure} &
        \begin{subfigure}[b]{0.097\linewidth}
            \includegraphics[width=\textwidth]{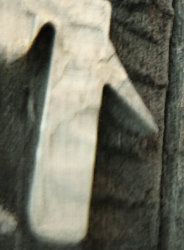}
            \caption{0.0730}
        \end{subfigure} &
        \begin{subfigure}[b]{0.097\linewidth}
            \includegraphics[width=\textwidth]{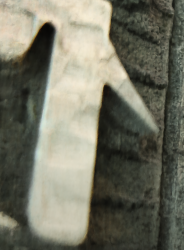}
            \caption{\seccol{0.0669}}
        \end{subfigure} &   
        \begin{subfigure}[b]{0.097\linewidth}
            \includegraphics[width=\textwidth]{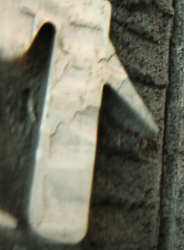}
            \caption{\tp{0.0599}}
        \end{subfigure} &  
        \begin{subfigure}[b]{0.097\linewidth}
            \includegraphics[width=\textwidth]{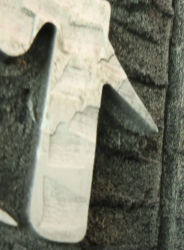}
            \caption{}
        \end{subfigure} \\

        % \hline
    \end{tabular}
    \caption{Visual comparisons of different radiance field modeling methods. The first two rows are synthetic scenes with motion and defocus blur. The last two rows are real scenes with motion and defocus blur. For real scenes, the ground truth is only for reference as its exposure is inconsistent with source views. Db-Ten.RF and Db-NeRF are shorthands for Deblur-TensoRF and Deblur-NeRF. We color code \tp{best} and \seccol{second best} performances.}
    \label{tab:visualization}
    \vspace{-1.5em}
\end{figure*}

\noindent\textbf{Ablation Study.} We compare three versions of designs to demonstrate the effectiveness of different components:

\begin{itemize}
    % \item DeblurTensoRF: Deblur-NeRF's rendering pipeline, i.e. NeRF, is replaced with TensoRF; we use 160 samples per ray for tracing. Details in supplemental material.
    \item $\textrm{PDRF}^{\textrm{single}}$-5: We use only the first stage, i.e. Eq. (4), to model blur without ray features, kernel size $P=5$.
    \item PDRF-5: PDRF with kernel size $P=5$.
    \item PDRF-10: PDRF with kernel size $P=10$.
\end{itemize}

As shown in Table 1, $\textrm{PDRF}^{\textrm{single}}$-5 is much faster than Deblur-NeRF and on par in performance, demonstrating the effectiveness of CRR and FVR; however, PDRF-5 outperforms $\textrm{PDRF}^{\textrm{single}}$-5 by introducing a second refinement stage in PBE. PDRF-5 is not significantly slower than $\textrm{PDRF}^{\textrm{single}}$-5, as CRR can obtain ray features very quickly. Finally, PDRF-10 uses a more powerful blur model by doubling the kernel size; this leads to significant performance improvement but also slower training time. An interesting direction for further optimization is to first distinguish the blurry and non-blurry regions, e.g. as shown in Fig. 3, and focus more modeling power on blurry regions. This also has applications to object motion modeling in NeRF, as oftentimes only some objects in the scene move when all observations are modeled for deformation. We demonstrate PDRF's application to \emph{object motion modeling} in the supplementary section.

\noindent\textbf{Quantitative and Visual Evaluation.}
We quantitatively and visually compare PDRF against various radiance field modeling methods in Table. 1 and Fig. 4. NeRF~\cite{DBLP:conf/eccv/MildenhallSTBRN20}, Plenoxel~\cite{fridovich2022plenoxels}, and TensoRF~\cite{tensorf} are SoTA methods that do not explicitly account for blur degradation. Plenoxel and TensoRF are significantly faster than NeRF; however, these methods often do not outperform NeRF when the inputs are blurry. In fact, TensoRF performs significantly worse and sometimes reaches collapsed solutions, e.g. with the Pool scene, where scene density is all concentrated near the camera origins. This is due to the limited capacity in TensoRF's low-rank density representation, which has more noise given blurry observations. Furthermore, the noise leads to less efficient voxel pruning and longer optimization time. Plenoxel performs better than TensoRF, but takes more memory at 2 GB per scene compared to TensoRF's 200 MB and PDRF's 160 MB features. We note that both Plenoxel and Instant-NGP~\cite{mueller2022instant} use customized CUDA kernels for acceleration and are difficult to be analyzed or modified. In particular, Instant-NGP does not have baselines for real forward-facing scenes like LLFF~\cite{DBLP:journals/tog/MildenhallSCKRN19}. We also tested on DVGO~\cite{SunSC22}; however, DVGO yields collapsed solutions for all scenes.

We then look into radiance field modeling methods that take blur into consideration. One approach is to first apply image-based or video-based deblurring methods on source views, as is done by MPR~\cite{DBLP:conf/cvpr/ZamirA0HK0021} and KPAC~\cite{9709995}. While this approach leads to some improvements, denoted as MPR+NeRF and KPAC+NeRF in Table 1, it does not explore 3D redundancy from different source views. Deblur-NeRF~\cite{ma2022deblur} uses a pixel-coordinate-based blur model before radiance field optimization and does not rely on external training data. In comparison, it outperforms all previous methods; however, Deblur-NeRF is also the most computationally demanding method and requires twenty hours to reach convergence. We create another baseline by replacing NeRF with TensoRF in Deblur-NeRF, which we named Deblur-TensoRF. Deblur-TensoRF is significantly faster than Deblur-NeRF; however, its performance is also notably worse. We contribute this again to TensoRF's compressed density representation. We also note that Deblur-TensoRF does not use voxel pruning. To demonstrate the incompatibility of voxel pruning with dynamic ray generation, we add a $\textrm{Deblur-TensoRF}^{\textrm{pru}}$ baseline in Table 1, which prunes low density voxels during training and results in significantly worse performances than Deblur-TensoRF. Our CRR+FVR design uses a combination of explicit and implicit representation to estimate density, and is both faster and more performant than Deblur-TensoRF; specifically, PDRF-5 is 15 times faster than Deblur-NeRF and takes only 80 minutes to converge. This is even faster than Deblur-TensoRF, partially due to the more efficient sampling scheme. We observe that PDRF's PBE can effectively model blur by using feature from the underlying radiance field, especially for defocus blur. This is likely because defocus blur can be much better modeled with 3D context, e.g. scene depth. Camera motion blur is less sensitive to 3D geometry of the scene; however, improvements can still be observed.

Visually, PDRF can recover very fine details both on synthesized and real scenes despite the blurry observations from neighboring source views. As shown in Fig. 4, Deblur-NeRF fails to recover details. While Deblur-TensoRF benefits from the additional blur modeling, it suffers from more noise. This can also be observed when comparing TensoRF to Plenoxel and NeRF. For more visualizations on scene density, please refer to the supplementary material section.

\vspace{-0.2em}
\section{Conclusion}
In this work, we present PDRF, a novel radiance field modeling method that addresses camera motion and defocus blur in observations. PDRF is both significantly faster and more performant than previous deblurring methods. Specifically, PDRF proposes PBE, which progressively updates the blur model by incorporating information from the underlying radiance field; as such, PBE can accurately account for the observed blur. To support PBE and accelerate radiance field optimization, PDRF proposes a rendering pipeline that consists of CRR and FVR. CRR is an efficient renderer that predicts ray color based on aggregated ray features. With its lightweight design, CRR provides ray feature estimation to PBE and density estimation to FVR. FVR uses importance sampling to determine efficient voxel samples along the ray, which are then rendered by a larger network for their color and density. Both CRR and FVR use explicit feature representation to help accelerate their convergence rate. We perform extensive experiments and show that our method leads to significantly better deblurring results, especially on defocus blur. PDRF is about 15X faster than Deblur-NeRF, which is the previous SoTA. In the future, we hope to investigate more efficient ways to increase the blur model capacity without sacrificing speed.% and introduce features to model other multi-view inconsistencies, e.g. object motion.

{\small
\bibliography{main}
}
% \newpage
% \input{supplementary}

\end{document}

\maketitle

\appendix
\section{Quantitative Evaluation on Real Scenes}

% \captionsetup[subfigure]{labelformat=empty}
% \newcommand{\tp}[1]{\colorbox{LimeGreen}{#1}}

\begin{table}[!htb]
    \setlength{\tabcolsep}{3pt}
    \centering
    \begin{tabular}[b]{c}
        \begin{tabular}{l|ccc}
        % \multicolumn{4}{c|}{} & \multicolumn{4}{c}\\
        \toprule
           \textsl{Camera Motion}& {PSNR} & {SSIM} & {LPIPS} \\
        \midrule
        NeRF & 22.95 & 0.6333 & 0.3742 \\
        MPR+NeRF & 23.38 & 0.6655 & 0.3140 \\
        Deblur-NeRF & 25.65 & 0.7586 & 0.1818 \\
        PDRF & 25.83 & 0.7642 & 0.2032 \\
        \midrule
        \textsl{Defocus} & {PSNR} & {SSIM} & {LPIPS} \\
        \midrule
        NeRF & 22.53 & 0.6627 & 0.2480\\
        KPAC+NeRF & 23.04 & 0.6917 & 0.1847\\
        Deblur-NeRF & 23.47 & 0.7244 & 0.1220\\ 
        PDRF & 23.84 & 0.7355 & 0.1319\\
        \bottomrule
        \end{tabular}\\
    \end{tabular}        
    \caption{Quantitative comparisons of radiance field modeling methods on twenty real scenes, ten are affected by camera motion blur and another ten are affected by defocus blur. The average statistics are reported for each blur type.}
    \label{tab:results_real}
\end{table}

We report the quantitative metrics on twenty real scenes from the Deblur-NeRF~\cite{ma2022deblur} dataset, as shown in Table 1. We note that it is difficult to obtain ground truth on real blurry images. The metrics are computed between synthesized novel views and references, which are acquired with different exposures, e.g. to ameliorate defocus blur. Furthermore, the calibration from COLMAP~\cite{DBLP:conf/cvpr/SchonbergerF16,DBLP:conf/eccv/SchonbergerZFP16} is likely not accurate due to blur. As works so far have only considered the effects of blur on radiance field, an interesting direction is to account for blur during pose estimation, e.g.~\cite{DBLP:conf/iccv/ParkL17} and radiance field optimization. In Table 1, we find that PDRF outperforms Deblur-NeRF on conventional metrics, i.e. PSNR and SSIM, while worse on LPIPS, which is based on a pre-trained VGG~\cite{DBLP:journals/corr/SimonyanZ14a}. One of the reasons that may account for the overall worse LPIPS score, despite better PSNR/SSIM, is the TV loss in PDRF during optimization. In areas that do not have sufficient information, e.g. near image boundaries, a TV loss optimizes for smoothness; such smoothness often leads to worse perceptual scores. 

\section{Object Motion Modeling}

% \vspace{-0.5em}
\begin{figure}[!htb]
    \setlength{\abovecaptionskip}{3pt}
    \setlength{\tabcolsep}{2pt}
    \begin{tabular}[b]{cc}
        \begin{subfigure}[b]{.48\linewidth}
            \includegraphics[width=\textwidth]{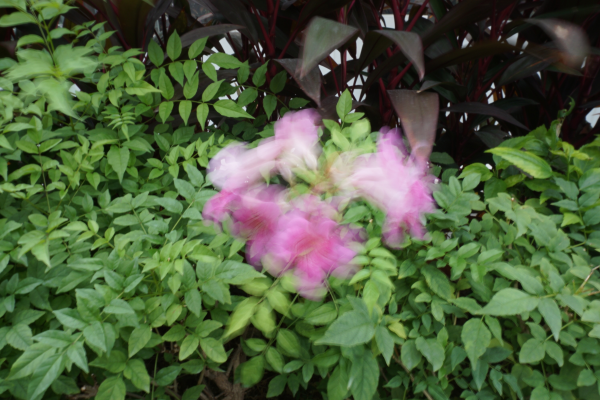}
            \caption{Object motion in src. views}
            % \label{img:objmo}
        \end{subfigure} &
        \begin{subfigure}[b]{.48\linewidth}
            \includegraphics[width=\textwidth]{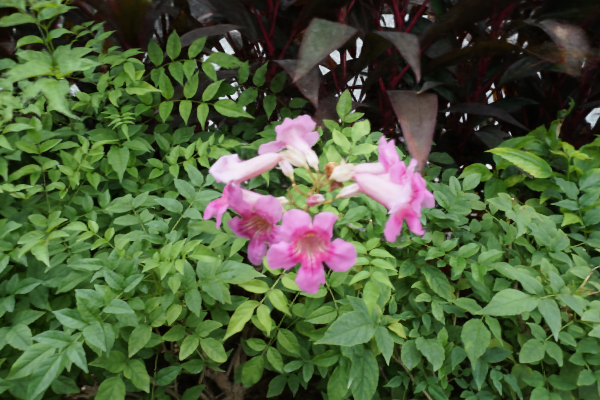}
            \caption{PDRF reconstruction}
            % \label{pdrf_objmo}
        \end{subfigure} \\
        \begin{subfigure}[b]{.48\linewidth}
            \includegraphics[width=\textwidth]{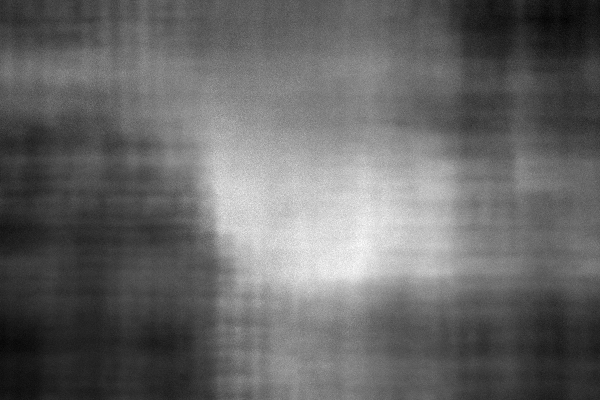}
            \caption{$Var(\Delta x^{*}_i)$, Ma et al.}
            % \label{img:DSKobjmo}
        \end{subfigure} &
        \begin{subfigure}[b]{.48\linewidth}
            \includegraphics[width=\textwidth]{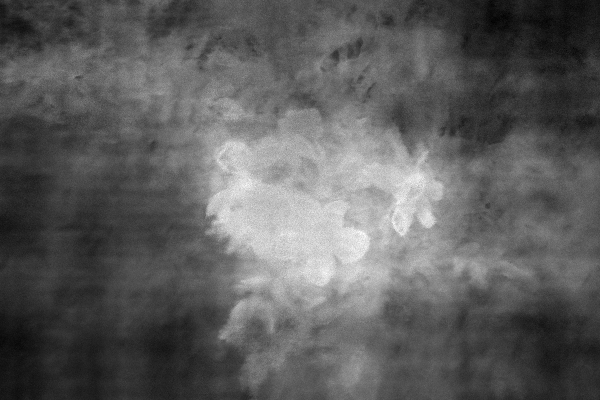}
            \caption{$Var(\Delta x_i)$, PBE}
            % \label{img:PREobjmo}
        \end{subfigure} \\
    \end{tabular}
    \caption{Visualization of object motion in (a), the foreground flower is in motion. We visualize the variance of predicted $\Delta x_i$, where higher variance correlates with more motion. Our PBE module (d) can clearly distinguish foreground details despite their proximity with background pixels; (c) can only coarsely distinguish fore/background.}
    % \label{fig:objmo}
    % \vspace{-1em}
\end{figure}

While PDRF is not specifically designed for object motion in source views, it can be straightforward used to account for this situation similar to Deblur-NeRF. We demonstrate an instance of this use case in Fig. 1, where the flower is clearly moving while the background is relatively stable. Fig. 1b shows that PDRF can indeed recover a motion-free radiance field. Furthermore, we observe similar improvements to detail modeling in Fig. 1c and 1d, when comparing rays estimated from the coarse and fine stage. In this scene, the background leaves also move slightly due to wind. In cases where motion and blur are more localized, interesting opportunities exist to explore the a sparsity in multi-view inconsistency to improve performance, while minimally increasing compute complexity.

\section{Additional Visualizations}

We provide additional visualizations on ray desntiy estimation, as shown in Fig. 2. 

\newpage
{\small
\bibliography{aaai23}
}